\documentclass[letterpaper]{article} % DO NOT CHANGE THIS
\usepackage[draft]{aaai25}  % DO NOT CHANGE THIS
\usepackage{times}  % DO NOT CHANGE THIS
\usepackage{helvet}  % DO NOT CHANGE THIS
\usepackage{courier}  % DO NOT CHANGE THIS
\usepackage[hyphens]{url}  % DO NOT CHANGE THIS
\usepackage{graphicx} % DO NOT CHANGE THIS
\usepackage{natbib}  % DO NOT CHANGE THIS AND DO NOT ADD ANY OPTIONS TO IT
\usepackage{caption} % DO NOT CHANGE THIS AND DO NOT ADD ANY OPTIONS TO IT
\usepackage{booktabs}
\usepackage{multirow}
\usepackage{subcaption}
\usepackage{amsmath}
\usepackage{xspace}
\usepackage{float} % For the H option
\usepackage[dvipsnames]{xcolor}
\usepackage{tikz}
\usepackage{array}
\usepackage{colortbl} % For coloring cells
\usepackage{float} % For the H option
\usepackage{adjustbox}
\usepackage{arydshln} % 在导言区添加这个包以支持虚线
\usepackage{pifont}
\definecolor{darkyellow}{RGB}{255, 217, 102}  % 这里的 RGB 值是姜黄色
\definecolor{darkyellow1}{RGB}{255, 230, 153}  % 这里的 RGB 值是姜黄色
\definecolor{darkorange}{RGB}{244, 177, 131}  % 这里的 RGB 值是姜黄色
\definecolor{darkblue}{RGB}{68, 114, 196}  % 这里的 RGB 值是姜黄色
\definecolor{darkgray}{RGB}{214, 220, 229}  % 这里的 RGB 值是姜黄色
\definecolor{darkgreen}{RGB}{169, 209, 142}  % 这里的 RGB 值是姜黄色

\usepackage{natbib}  % DO NOT CHANGE THIS AND DO NOT ADD ANY OPTIONS TO IT
\usepackage{caption} % DO NOT CHANGE THIS AND DO NOT ADD ANY OPTIONS TO IT
\usepackage{algorithm}
\usepackage{algorithmic}

\usepackage{newfloat}
\usepackage{listings}
\DeclareCaptionStyle{ruled}{labelfont=normalfont,labelsep=colon,strut=off} % DO NOT CHANGE THIS
\floatstyle{ruled}
\newfloat{listing}{tb}{lst}{}
\floatname{listing}{Listing}
\title{FedRGL: Robust Federated Graph Learning for Label Noise}
\author{
   De Li,
   Haodong Qian,
    Qiyu Li,
    Zhou Tan,
    Zemin Gan,
    Jinyan Wang,
    Xianxian Li
}
\affiliations{
    \textsuperscript{\rm 1}Key Lab of Education Blockchain and Intelligent Technology, Ministry of Education, Guangxi Normal University\\
     \textsuperscript{\rm 2}Guangxi Key Lab of Multi-source Information Mining $\&$ Security, Guangxi Normal University\\
    \textsuperscript{\rm 3}School  of Computer Science and Engineering, Guangxi Normal University
    
}

\usepackage{bibentry}
\begin{document}

\maketitle
\begin{abstract}
Federated Graph Learning (FGL) is a distributed machine learning paradigm based on graph neural networks, enabling secure and collaborative modeling of local graph data among clients. However, label noise can degrade the global model's generalization performance. Existing federated label noise learning methods, primarily focused on computer vision, often yield suboptimal results when applied to FGL. To address this, we propose a robust federated graph learning method with label noise, termed FedRGL. FedRGL introduces dual-perspective consistency noise node filtering, leveraging both the global model and subgraph structure under class-aware dynamic thresholds. To enhance client-side training, we incorporate graph contrastive learning, which improves encoder robustness and assigns high-confidence pseudo-labels to noisy nodes. Additionally, we measure model quality via predictive entropy of unlabeled nodes, enabling adaptive robust aggregation of the global model. Comparative experiments on multiple real-world graph datasets show that FedRGL outperforms 12 baseline methods across various noise rates, types, and numbers of clients.
\end{abstract}

\section{Introduction}
Graphs are widely used in modeling complex systems due to their ability to visually represent relational information between different entities \cite{bang2023biomedical,yang2023dgrec,guo2020deep}. Graph neural networks (\textbf{GNNs}) \cite{defferrard2016convolutional}, as a promising method for graph information mining, have achieved excellent performance in node-level \cite{zhu2020beyond,fu2023hyperbolic}, edge-level \cite{cai2021line,yuan2024dynamic}, and graph-level \cite{sun2022graph,ju2023unsupervised} downstream tasks. However, most of the existing GNNs training based on centralized data storage is not applicable to real-world scenarios due to data privacy protection as well as copyright constraints. Therefore, Federated Graph Learning (\textbf{FGL}) \cite{xie2021federated,huang2023federated,tan2023federated,li2023fedgta,wan2024federated,zhu2024fedtad}, which combines graph learning and federated learning, has been proposed to enable joint modeling between clients under the protection of private graph data not going out of the local area.
\begin{figure}[!t]
    \centering
    \includegraphics[width=0.46\textwidth]{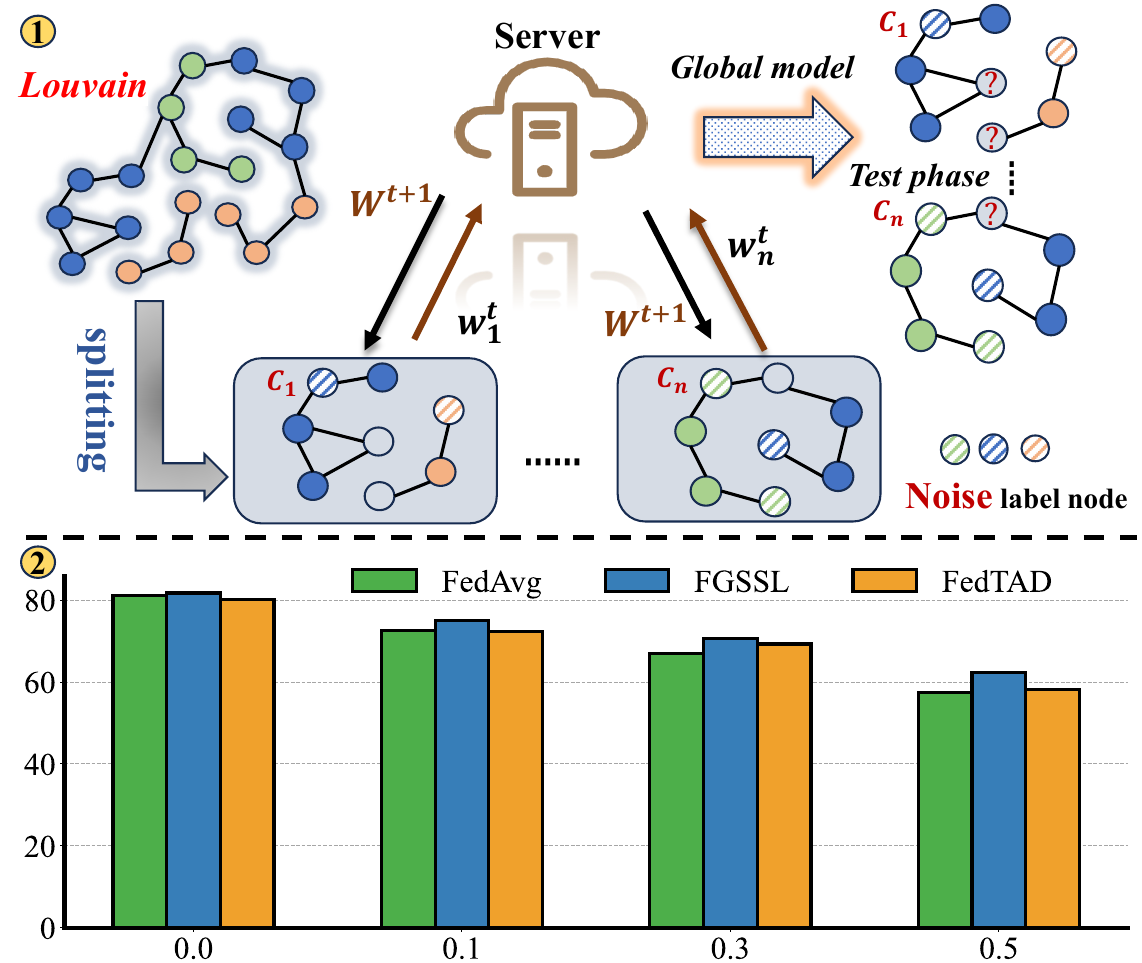}
    \caption{\protect\tikz[baseline=-0.5ex]\protect\node[fill=darkyellow,draw,circle,inner sep=0.05em, scale=0.75]{\textbf{1}}; Example of Federated Graph Learning with noisy labels. Subgraph data is divided among clients using the Louvain algorithm \cite{blondel2008fast}, with varying noise levels. The collaboratively trained global model is tested for final performance. 
    \protect\tikz[baseline=-0.5ex]\protect\node[fill=darkblue,draw,circle,inner sep=0.25em]{};, 
    \protect\tikz[baseline=-0.5ex]\protect\node[fill=darkgreen,draw,circle,inner sep=0.25em]{};, and 
    \protect\tikz[baseline=-0.5ex]\protect\node[fill=darkorange,draw,circle,inner sep=0.25em]{}; 
    represent training nodes, while 
    \protect\tikz[baseline=-0.5ex]\protect\node[fill=darkgray,draw,circle,inner sep=0.25em]{}; 
    represents test nodes.
    \protect\tikz[baseline=-0.5ex]\protect\node[fill=darkyellow,draw,circle,inner sep=0.05em, scale=0.75]{\textbf{2}}; 
    Testing accuracy of three FGL algorithms on Cora under \textbf{uniform} label noise at different rates (\textit{10 clients}), showing lack of robustness to label noise.}
    \label{G1}
\end{figure}

% \begin{figure*}
%     \centering
%     \begin{subfigure}{0.24\textwidth}
%         \centering
%         \includegraphics[width=\textwidth]{G2.pdf}
%         \caption{FNL Performance}
%         \label{g21}
%     \end{subfigure}
%     \hspace{0.00\textwidth}
%     \begin{subfigure}{0.243\textwidth}
%         \centering
%         \includegraphics[width=\textwidth]{G3.pdf}
%         \caption{GNL Performance}
%         \label{g22}
%     \end{subfigure}
%     \hspace{0.00\textwidth}
%     \begin{subfigure}{0.24\textwidth}
%         \centering
%         \includegraphics[width=\textwidth]{G4.pdf}
%         \caption{Label Distributions}
%         \label{g23}
%     \end{subfigure}
%     \hspace{0.00\textwidth}
%     \begin{subfigure}{0.24\textwidth}
%         \centering
%         \includegraphics[width=\textwidth]{G5.pdf}
%         \caption{Performance Comparison}
%         \label{g24}
%     \end{subfigure}
%     \caption{\subref{g21} Global test accuracy of different FNL methods on Cora with 10 clients at 0.3 noise rate. \subref{g22} Global test accuracy of different GNL methods on Cora with 10 clients under varying noise rates. \subref{g23} Class label distribution across 10 clients on Cora. \subref{g24} Test accuracy comparison of 12 methods on six real graph datasets under different noise conditions.}
%     \label{G2}
% \end{figure*}
Current academic research on FGL primarily focuses on optimizing the performance of global or personalized models under Non-independent and identically distributed (Non-iid) graph data. For example,  FedTAD \cite{zhu2024fedtad} proposes topology-aware knowledge distillation to transfer knowledge from local models to the server to obtain an optimal global model. FedStar \cite{tan2023federated} decouples structural encoding and feature encoding to learn invariant knowledge from heterogeneous domains for personalized models on clients.  However, existing FGL approaches assume that local graph data is jointly trained under the premise of completely clean labels, neglecting the impact of noisy labels in client graph data on the model optimization process, as illustrated in Fig. \ref{G1}. Therefore, this paper further explores and enhances the robustness of Federated Graph Learning under the influence of noisy labels.

Existing efforts to address the noisy label problem in federated learning (FNL) \cite{xu2022fedcorr,fang2022robust,wu2023fednoro,li2024feddiv,lu2024federated} are largely focused on computer vision (CV), but their direct application to federated graph learning (FGL) often fails to yield performance gains. This is due to the structural heterogeneity among federated subgraphs (Fig. \ref{G1}) and class imbalances within local subgraphs \cite{zhu2024fedtad}, which complicate the identification and management of noisy nodes (\textit{See the Appendix for further validation}). Current FNL methods employ uniform thresholds for sample selection (e.g., FedCorr \cite{xu2022fedcorr}), modify local objective functions (e.g., RHFL \cite{fang2022robust}, FedNoro \cite{wu2023fednoro}), or optimize global models via negative knowledge distillation (e.g., FedNed \cite{lu2024federated}). However, these methods lack fine-grained noisy node processing within local subgraphs, leading to limited performance gains. Moreover, current research on the impact of noisy labels in graphs mainly targets centralized scenarios, including graph semi-supervised node classification \cite{dai2021nrgnn,qian2023robust,liu2023multi,chen2023erase,xia2023gnn,li2024contrastive}, graph classification \cite{yin2023omg}, and graph transfer learning \cite{yuan2023alex}. Existing graph noise learning (\textbf{GNL}) methods typically assume complete graph structure information for noise mitigation. However, in federated graph learning, clients only possess local subgraph information, making the incomplete graph structure and limited training data significant challenges for effectively mitigating noise in FGL.

In this paper, we propose FedRGL, a Federated Graph Learning  method that combines global model information and subgraph structural information to mitigate the impact of noisy labels on the global model. \textbf{On the client side}, unlike previous methods that set a uniform threshold and use the Gaussian Mixture Model (GMM) to fit sample losses \cite{xu2022fedcorr,li2024feddiv}, FedRGL addresses the heterogeneity issue of class distribution in local clients. This heterogeneity results in different learning rates of class knowledge for local GNN models \cite{zhu2024fedtad}. Therefore, FedRGL uses the global model and a corrected local subgraph label propagation algorithm to calculate sample loss values separately and achieves precise selection of noisy nodes under class-aware loss dynamic threshold constraints. To utilize the useful information from noisy nodes, FedRGL introduces a graph contrastive method in local GNN training. This not only enhances the noise robustness of the encoder but also employs graph contrastive augmentation to assign high-confidence pseudo-labels to noisy nodes, further increasing the supervisory information for model training. \textbf{On the server side}, due to varying noise levels among clients, the quality of client-uploaded models must be considered during global aggregation. This paper leverages transductive learning for local subgraphs \footnote{Transductive learning assumes that all node feature and structure information is known during training, but only the labels of training nodes are available} for local subgraphs., measuring model quality via prediction entropy on unlabeled nodes before uploading to the server, ensuring robust global model aggregation.  

\textbf{Our contributions.} (1) \underline{\textit{Novel Research}}. To the best of our knowledge, this is the first study to address label noise robustness in federated graph learning, providing new insights on enhancing the robustness of the federated global model under local subgraph heterogeneity and label noise. (2) \underline{\textit{New Method}}. We propose FedRGL, a label noise learning approach that integrates global model knowledge with local subgraph structural knowledge. By accounting for class distribution and local subgraph training characteristics, FedRGL enables precise noisy node selection and robust global model aggregation. (3) \underline{\textit{State-of-the-art Performance}}. FedRGL achieves superior test accuracy across multiple real-world graph datasets, consistently outperforming various baseline methods under different noise types, noise rates, and client numbers.

\section{Related Work}
\textbf{Federated Graph Learning.} Existing Federated Graph Learning (FGL) methods can be broadly classified into two types based on the type of graph task: graph FL \cite{xie2021federated,tan2023federated} and subgraph FL \cite{zhang2021subgraph,huang2023federated,zhu2024fedtad,baek2023personalized}. In graph FL, each client owns a set of graph data and designs personalized graph mining methods to accomplish graph-level classification tasks, such as FedStar approach \cite{tan2023federated} of using a feature and structure decoupled encoder to learn invariant structural knowledge between clients alongside personalized knowledge from local data. In contrast, subgraph FL involves each client holding only partial subgraph knowledge of a complete graph, with node-level classification tasks being completed by collaboratively training a global model or personalized local models. For example, FGSSL \cite{huang2023federated} enhances learning of graph nodes and structural information through supervised graph contrastive learning and relational distillation. This paper studies the generalization performance of global models from a robustness perspective, focusing on the dual challenges posed by inter-client graph structure heterogeneity and intra-client label noise.\\

\noindent\textbf{Federated Learning with label noise.} Current Federated Learning with label noise (FNL) research is mainly divided into methods based on loss-level and sample-level approaches. \protect\tikz[baseline=-0.5ex]\protect\node[fill=black,draw,circle,inner sep=0.05em, scale=0.75]{\textcolor{white}{\textbf{1}}}; \textit{Loss-level}. RHFL \cite{fang2022robust} utilizes a symmetric robust loss and KL divergence constrained on a public dataset to train personalized robust models. FedNoro \cite{wu2023fednoro} employs knowledge distillation and a distance-aware aggregation function to jointly update the noise-robust global model. \protect\tikz[baseline=-0.5ex]\protect\node[fill=black,draw,circle,inner sep=0.05em, scale=0.75]{\textcolor{white}{\textbf{2}}}; \textit{Sample-level}. FedCorr \cite{xu2022fedcorr}uses a GMM algorithm to filter out noisy clients and noisy samples, and re-labels the selected samples using the global model. Compared to the use of local GMMs to filter noisy samples, FedDiv \cite{li2024feddiv} collaborates with all clients to build global GMM parameters for the precise selection of noisy samples. FedFixer \cite{ji2024fedfixer} introduces a dual collaborative network of personalized and global models to alternately select noisy samples.\\

\noindent\textbf{Graph Neural Networks with Label Noise.} Existing research on noisy labels in graph neural networks primarily focuses on centralized scenarios, which can be broadly categorized into three types: graph structure augmentation level, graph contrast level, and graph structure propagation level. \textit{Graph structure augmentation level}: NRGNN \cite{dai2021nrgnn} and RTGNN \cite{qian2023robust} both enhance the propagation of graph information by linking labeled and unlabeled nodes, with the difference that the latter employs a dual-network structure to further prevent error accumulation. \textit{Graph contrast level}: Leveraging the robustness of unsupervised graph contrastive methods to label noise, CRGNN \cite{li2024contrastive} improves model test performance by using dynamic loss and cross-space consistency, while CGNN \cite{yuan2023learning} corrects labels using neighbor label information. \textit{Graph structure propagation level}: Benefiting from the efficiency of label propagation algorithms, GNN-cleaner \cite{xia2023gnn} and R\textsuperscript{2}LP \cite{2023arXiv231016560C} address noise in semi-supervised node classification when a certain amount of clean training node label information is available. Earse \cite{chen2023erase} further relaxes the constraint of clean sample information by learning error-tolerant node representations using prototype pseudo-labels and structure-propagated pseudo-labels.\\

\noindent\textbf{Connections and Differences.} Unlike existing FNL methods targeting the CV domain, this study focuses on the more complex problem of learning with label noise in federated subgraphs, where the structural heterogeneity of subgraphs and label noise across clients exacerbate the difficulty of filtering noisy nodes, making it challenging to directly apply existing FNL methods. Moreover, FedRGL leverages the label propagation algorithm to filter noisy nodes within clients, but it differs from existing methods in three key aspects. 
\textbf{First}, our proposed method does not require a prior condition of clean labels, whereas both GNN-cleaner and R\textsuperscript{2}LP rely on the assistance of clean label priors. 
\textbf{Second}, before label propagation, FedRGL uses a robust global model to assess and correct the initial node labels, whereas ERASE directly utilizes the initial node label information. 
\textbf{Lastly}, the existing three methods directly utilize the one-hot label information after label propagation for label correction, while FedRGL computes the loss using the propagated soft labels (e.g., probability distributions) along with the original labels, providing a structural perspective for noisy node filtering.
\section{Methodology}
\subsection{\textbf{Preliminaries}}

In subgraph FGL, there exist \( M \) clients that collaboratively train a global model with a central server. The \( m \)-th client possesses part of the global graph \( G_g = (V_g, A_g, X_g, Y_g) \), represented as a subgraph \( G_m = (V_m, A_m, X_m, Y_m) \), where \( V_m \) contains training nodes \( V_m^{T_r} \), validation nodes \( V_m^{V_a} \), and test nodes \( V_m^{T_e} \). Each node \( v_i \in V_m \) has a feature vector \( x_i^m(x_i^m \subseteq X_m) \) and a label \( y_i^m(y_i^m \subseteq Y_m) \), where \( y_i^m \) may be noisy. In this paper, we assume that the label noise rate among clients varies and is obtained through uniform sampling \( U(\eta^l, \eta^u) \) \cite{xu2022fedcorr}. FedAvg, as the baseline model, serves as the foundation for the FedRGL method proposed in this paper. Each client performs local GNN model training by minimizing the loss function \( L(G_m) \), and then uploads the model parameters \( w_m \) along with the number of nodes \( |V_m| \) to the central server. The central server performs a federated weighted average to obtain the global model \(W\). Finally, after multiple rounds of collaborative training, the global model will be sent to each client for testing tasks.
\begin{figure*}
    \centering
    \begin{subfigure}{1\textwidth}
        \centering
        \includegraphics[width=\textwidth]{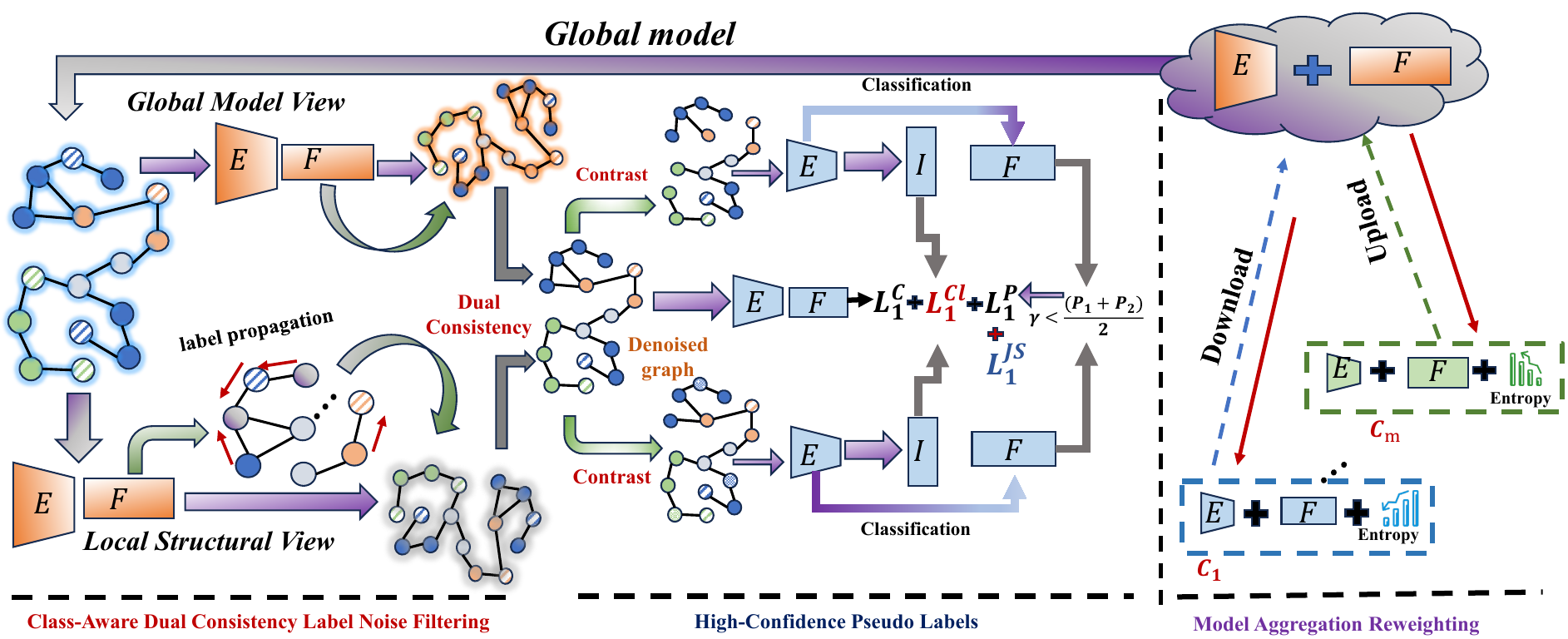}
        \label{fig:sub3}
    \end{subfigure}
    \caption{Overall framework diagram of the FedRGL method.}
    \label{G3}
\end{figure*}
\subsection{\textbf{FedRGL Method}}
\textbf{Overview}. The FedRGL method, as shown in Fig. \ref{G3}, operates with distinct client-side and server-side processes. \textit{Client-side}, each client computes the training loss using both the global model and local structure, filtering noisy nodes with a class-aware dynamic threshold (Class-Aware Dual Consistency Label Noise Filtering). A graph contrastive approach then assigns high-confidence pseudo-labels to the filtered nodes, improving the quality of the training data. After local training, the clients calculate and upload the prediction entropy of unlabeled nodes to the server. \textit{Server-side}, the server performs aggregation reweighting based on the clients' prediction entropy to achieve a robust global model, which is then broadcasted back to the clients for the next training round. This process repeats until the training is complete.

\noindent\textbf{Class-Aware Dual Consistency Label Noise Filtering}\\
In subgraph FL, structural heterogeneity among clients complicates noisy node identification and management. To address this, we propose a dual-consistency label noise filtering method that considers both global model and local structural views. \underline{\textbf{Global Model View}}. Before the $t$-th training round, the $m$-th client uses the global model $W^t$ to calculate the cross-entropy loss \( L(G_m) \) for each training node as $\langle \mathcal{L}{v_1}, \mathcal{L}{v_2}, \ldots, \mathcal{L}{v_k} \rangle$. Unlike methods that use a GMM to fit loss values and set a unified clean probability threshold, our approach considers class distribution differences and intra-class homophily in client subgraphs. As different node categories are learned at varying rates, a unified global threshold is ineffective. Instead, we propose class-aware dynamic thresholds, calculating the mean $t_{m}^c$ and standard deviation $\sigma_{m}^c$ of the cross-entropy loss for each class:
\begin{equation}
V_m^c = \{v_i \mid y_i^m = c\}, t_{m}^c = \frac{1}{|V_m^c|} \sum_{v_i \in V_m^c} \mathcal{L}_{v_i}
\end{equation}
\begin{equation}
\sigma_{m}^c = \sqrt{\text{var}(\mathcal{L}_{v_i})}, v_i \in V_m^c
\end{equation}
\begin{equation}
\rho_m^c = t_{m}^c + \varphi_1 \sigma_{m}^c
\end{equation}
where $V_m^c$ represents the set of nodes in the $m$-th client that belong to class $c$, $\rho_m^c$ is the noise node filtering threshold for class $c$, and $\varphi_1$ is a hyperparameter. Then, according to the dynamic threshold set $\{\rho_m^1, \rho_m^2, \ldots, \rho_m^C\}$ of all classes in the $m$-th client, the clean sample node set $V_{m}^{C_1}$ under the global model view of the client can be obtained as:
\begin{equation}
V_{m}^{C_1} = \{v_i \mid y_i^m = c, \mathcal{L}_{v_i} < \rho_m^c\}
\end{equation}
\underline{\textbf{Local Structural View}}. We introduce the label propagation calculation after global model correction to further enhance the stability and accuracy of noise node filtering. Specifically, to avoid the influence of initial label noise in the local label propagation process, we use the global model $W^{t}$ to predict the label distribution $P_{m} = \text{Softmax}(f(V_{m}^{T_r}, W^{t}))$ for the training nodes of the $m$-th client at the $t$-th communication round. For the nodes $v_{m}^{Eq}$ whose predicted labels $\text{argmax}(P_{m})$ match their original labels $Y_{m}$, we retain their original labels during the label propagation process and encode them as one-hot. For the remaining training nodes $V_{m}^{Re} = V_{m}^{T_r} \setminus V_{m}^{Eq}$, the soft labels predicted by the global model are used for initialization.

To avoid the impact of non-training nodes, we employ the subgraph structure masking technique in Earse \cite{chen2023erase}, constructing a masking matrix $MM^{T}$, where $M \in \{0,1\}^{N}$, such that the label propagation process only transmits information within the training nodes. The adjacency matrix is denoted by $A_{m}' = A_{m} \odot MM^{T}$. By introducing a non-parametric label propagation algorithm for $k$ steps, we can obtain the class probabilities guided by structural information for each training node:
\begin{equation}
(\hat{Y}_{m}^{T_r})^{k} = \alpha(\hat{Y}_{m}^{T_r})^{k-1} + (1 - \alpha)\left(D^{-\frac{1}{2}}A_{m}'D^{-\frac{1}{2}}\right)(\hat{Y}_{m}^{T_r})^{k-1}
\end{equation}
where $D$ is the diagonal matrix, and $\alpha$ is the hyperparameter for label propagation. Particularly, instead of performing label noise processing with hard labels (\textit{i.e.,} one-hot vectors) after label propagation \cite{chen2023erase,xia2023gnn}, we obtain soft labels (\textit{i.e.,} class probability distribution) here. Then, using the soft labels after structural propagation and the original labels of the training node $V_{m}^{T_r}$ to compute the cross-entropy value $\langle \ell_{v_1}, \ell_{v_2}, \ldots, \ell_{v_k} \rangle$, and the set of dynamic thresholds for noise nodes lost by class for the $m$-th client can be obtained as $\{\mu_{m}^{1}, \mu_{m}^{2}, ..., \mu_{m}^{c}\}$, where $\mu_{m}^{c} = t_{m}^{c} + \varphi_{2} \sigma_{m}^c$. By calculating the dynamic threshold value for each class, we can obtain the clean sample set under the local structural view:
\begin{equation}
V_{m}^{C_2} = \{v_{i} | y_{m}^{i} = c, \ell_{v_{i}} < \mu_{m}^{c}\}
\end{equation}

Finally, through the intersection of the clean sample set $V_{m}^{C_1}$ under the global model view and the clean sample set $V_{m}^{C_2}$ under the local structural view, we can obtain the final clean sample set $V_{m}^{C} = V_{m}^{C_1} \cap V_{m}^{C_2}$ and the corresponding noisy node set $V_{m}^{N} = V_{m}^{T_r} \setminus V_{m}^{C}$. It is worth noting that to obtain a more optimal and stable performance, we set two hyperparameters $\varphi_{1}, \varphi_{2}$ for the dynamic threshold, and the experiment found that setting $\varphi_{1} = \varphi_{2}$ does not significantly degrade the model's generalization ability. \textbf{In addition}, our noise node selection is executed only once in each communication round, without performing multiple selections along with local epochs.

\noindent\textbf{High-Confidence Pseudo Labels}\\
To enhance supervision during local training in subgraph FL, we incorporate graph contrastive learning (GCL) into the local parameter optimization process. GCL not only improves the noise robustness of the local model's encoder but also assigns pseudo-labels to noisy nodes $V_m^{N}$ using dual-enhanced graph views, thereby increasing the labeling information for local training. Specifically, in the $t$-th round, the $m$-th client identifies clean nodes $V_m^{C}$ and noisy nodes $V_m^{N}$ through label noise filtering. We use edge drop and feature masking techniques from GRACE \cite{zhu2020deep} to construct two augmented views of the original graph $G_m$, denoted as $G_m^{1}$ and $G_m^{2}$, and obtain corresponding embeddings $Z_m^{1}$ and $Z_m^{2}$ via the local model's encoder $E_m$ and projection head $I_m$. Additionally, class predictions $p_m^{1}$ and $p_m^{2}$ are generated (without passing through $I_m$) using the classifier $F_m$. The noise robustness of the encoder is further enhanced in the embedding space through contrastive loss:
\begin{equation}
L_{m}^{Cl} = \frac{1}{2N}\left( L_{cl}\left(Z_m^{\text {1}}, Z_m^{\text {2}}\right) + L_{cl}\left(Z_m^{\text {2}}, Z_m^{\text {1}}\right)\right)
\end{equation}
\begin{equation}
L_{cl}(Z^1, Z^2) = \frac{\sum\limits_{i=1}^N \psi\left(Z^{1}_i, Z^{2}_i\right)}{\sum\limits_{j=1, j \neq i}^N \psi\left(Z^{1}_i, Z^{1}_j\right) + \sum\limits_{i=1}^N \psi\left(Z^{1}_i, Z^{2}_j\right)}
\end{equation}
where $\psi(a, b) = \exp(\text{\textbf{sim}}(a, b)/\tau)$, \textbf{sim} is the similarity function, and $\tau$ is the contrastive parameter set to 0.5 in this paper. For noisy nodes $V_m^{\text{N}}$, we use the class prediction results from the two augmented views and set a prediction confidence threshold $\gamma$ to obtain the corresponding pseudo-labels:
\begin{equation}
\tilde{Y}_m=\operatorname{Softmax}\left(\left(P_m^{\text {1}}+P_m^{\text {2}}\right) / 2\right)
\end{equation}
\begin{equation}
V_m^{\text {P}}=\left\{\left(v_m^i, y_m^i\right) \mid \max \left(y_m^i\right)>\gamma, v_m^i \in V_m^{{N}}\right\}
\end{equation}
where $y_m^i \in \tilde{Y}_m$. Therefore, the $m$-th client uses high-confidence pseudo-labels to increase the local supervision loss information $L_m^{\text{P}}$ as:
\begin{equation}
L_m^{\text {P}}=\frac{\sum\limits_{v_m^i \in V_m^{\text {p }}}\left(L_{c e}\left(\hat{P}_m^{\text {1},i }, \hat{y}_m^i\right)+L_{c e}\left(\hat{P}_m^{\text {2}, i}, \hat{y}_m^i\right)\right)}{2}
\end{equation}
where $L_{\text{ce}}$ is the cross-entropy loss, $\hat{P}_m^{\text{1},i} = \text{softmax}\left(P_m^{\text{1},i}\right)$, $\hat{y}_m^i = \text{argmax}\left({y}_m^i\right)$. To further enhance the stability of pseudo-labeled nodes during local training, we introduce the Jensen Shannon (JS) divergence for pseudo-labeled nodes $V_m^{\text{P}}$ augmented view predictions $G_m^{\text{1}}, G_m^{\text{2}}$ to ensure consistency with the predictions of the original graph $G_m$. Consequently, the updated loss function for the $m$-th client is expressed as:

\begin{equation}
L_m=L_m^{C}+\lambda_{C l} L_m^{C l}+\lambda_{\text {P }} L_m^{P }+\lambda_{\text{Js}} L_m^{J s}\left(\hat{P}_m, \hat{P}_m^{\text {1}}, \hat{P}_m^{\text {2}}\right)
\end{equation}
where $L_m^{C}$ is the cross-entropy loss for clean nodes $V_m^{\text{C}}$, and $L_m^{Js}$ is the JS divergence for pseudo-labeled nodes $V_m^{\text{P}}$. $\lambda_{C l}, \lambda_{\text{P}}, \lambda_{\text{Js}}$ are hyperparameters. \textbf{Notably}, the pseudo-labels obtained for noisy nodes will only be used in the current epoch and will not overwrite the original label information, in order to avoid accumulating errors.
\begin{table*}[t]
  \centering
  \begin{subtable}{\textwidth}
   \resizebox{\textwidth}{!}{
    \begin{tabular}{c|ccc|ccc|ccc}
        \toprule
        \rowcolor[rgb]{ .749,  .749,  .749} \textbf{Dataset} & \multicolumn{3}{c|}{\textbf{Cora (5 Clients)}} & \multicolumn{3}{c|}{\textbf{CiteSeer (5 Clients)}} & \multicolumn{3}{c}{\textbf{PubMed (10 Clients)}} \\
        \toprule
        \midrule
        Method & Normal & Uniform & Pair  & Normal & Uniform & Pair  & Normal & Uniform & Pair \\
        \midrule
        \midrule
        FedAvg & 81.04\textsubscript{$\pm$0.37} & 48.19\textsubscript{$\pm$0.34} & 55.80\textsubscript{$\pm$0.18} & 71.12\textsubscript{$\pm$0.44} & 38.73\textsubscript{$\pm$0.84} & 46.62\textsubscript{$\pm$0.79} & 85.86\textsubscript{$\pm$0.07} & 74.36\textsubscript{$\pm$0.45} & 71.21\textsubscript{$\pm$0.28} \\
        FedProx & 47.98\textsubscript{$\pm$0.42} & 47.98\textsubscript{$\pm$0.42} & 56.07\textsubscript{$\pm$0.55} & 70.83\textsubscript{$\pm$0.42} & 38.91\textsubscript{$\pm$0.72} & 46.65\textsubscript{$\pm$0.72} & 85.89\textsubscript{$\pm$0.02} & 74.36\textsubscript{$\pm$0.44} & 71.25\textsubscript{$\pm$0.27} \\
        FGSSL & 82.16\textsubscript{$\pm$0.21} & 51.81\textsubscript{$\pm$0.56} & 57.58\textsubscript{$\pm$0.85} & 71.88\textsubscript{$\pm$0.35} & 43.27\textsubscript{$\pm$0.88} & 49.78\textsubscript{$\pm$1.24} & \cellcolor[rgb]{ 1,  .902,  .6}86.44\textsubscript{$\pm$0.15} & 77.57\textsubscript{$\pm$0.11} & 73.07\textsubscript{$\pm$0.17} \\
        FedTAD & 81.82\textsubscript{$\pm$1.01} & 47.20\textsubscript{$\pm$1.26} & 55.40\textsubscript{$\pm$0.37} & 70.41\textsubscript{$\pm$1.09} & 39.35\textsubscript{$\pm$0.29} & 45.98\textsubscript{$\pm$0.80} & 85.65\textsubscript{$\pm$0.05} & 73.46\textsubscript{$\pm$0.66} & 72.11\textsubscript{$\pm$0.19} \\
        Co-teaching & 81.04\textsubscript{$\pm$0.67} & 60.11\textsubscript{$\pm$2.09} & 57.68\textsubscript{$\pm$5.36} & 70.61\textsubscript{$\pm$0.41} & 45.13\textsubscript{$\pm$1.54} & 53.17\textsubscript{$\pm$2.95} & 85.71\textsubscript{$\pm$0.09} & 58.32\textsubscript{$\pm$6.98} & 60.21\textsubscript{$\pm$7.64} \\
        FedCorr & 79.05\textsubscript{$\pm$0.82} & 50.14\textsubscript{$\pm$0.37} & 55.16\textsubscript{$\pm$0.97} & 71.42\textsubscript{$\pm$0.32} & 53.28\textsubscript{$\pm$1.39} & 51.00\textsubscript{$\pm$1.17} & 81.18\textsubscript{$\pm$0.61} & 75.04\textsubscript{$\pm$0.67} & 70.90\textsubscript{$\pm$2.41} \\
        FedNoro & 81.19\textsubscript{$\pm$0.26} & 45.54\textsubscript{$\pm$1.02} & 45.80\textsubscript{$\pm$2.52} & 69.99\textsubscript{$\pm$0.63} & 37.53\textsubscript{$\pm$0.64} & 43.71\textsubscript{$\pm$0.26} & 84.10\textsubscript{$\pm$0.16} & 72.37\textsubscript{$\pm$0.34} & 65.99\textsubscript{$\pm$0.04} \\
        FedNed & 80.68\textsubscript{$\pm$0.00} & 47.11\textsubscript{$\pm$0.04} & 56.85\textsubscript{$\pm$1.02} & 69.50\textsubscript{$\pm$0.71} & 37.45\textsubscript{$\pm$0.06} & 47.90\textsubscript{$\pm$0.07} & 85.76\textsubscript{$\pm$0.18} & 74.44\textsubscript{$\pm$0.63} & 70.83\textsubscript{$\pm$0.03} \\
        RHFL  & 80.17\textsubscript{$\pm$0.47} & 49.92\textsubscript{$\pm$0.22} & 53.36\textsubscript{$\pm$0.99} & 66.05\textsubscript{$\pm$1.11} & 39.80\textsubscript{$\pm$1.04} & 47.88\textsubscript{$\pm$0.39} & 84.25\textsubscript{$\pm$0.07} & 61.87\textsubscript{$\pm$0.81} & 59.70\textsubscript{$\pm$0.49} \\
        CRGNN & \cellcolor[rgb]{ 1,  .902,  .6}83.09\textsubscript{$\pm$0.87} & 61.67\textsubscript{$\pm$0.80} & 63.24\textsubscript{$\pm$0.74} & 71.06\textsubscript{$\pm$0.59} & 48.09\textsubscript{$\pm$1.00} & 52.21\textsubscript{$\pm$1.77} & 84.70\textsubscript{$\pm$0.11} & 78.28\textsubscript{$\pm$0.31} & 73.46\textsubscript{$\pm$0.44} \\
        RTGNN & 80.65\textsubscript{$\pm$0.24} & 49.40\textsubscript{$\pm$0.19} & 57.50\textsubscript{$\pm$0.79} & 71.08\textsubscript{$\pm$0.43} & 39.45\textsubscript{$\pm$0.39} & 48.13\textsubscript{$\pm$0.54} & 85.09\textsubscript{$\pm$0.06} & 77.80\textsubscript{$\pm$0.76} & 73.04\textsubscript{$\pm$0.36} \\
        ERASE & 71.40\textsubscript{$\pm$0.16} & 51.81\textsubscript{$\pm$0.63} & 53.34\textsubscript{$\pm$0.82} & 63.42\textsubscript{$\pm$0.80} & 44.68\textsubscript{$\pm$0.75} & 47.76\textsubscript{$\pm$3.90} & 76.98\textsubscript{$\pm$0.36} & 67.00\textsubscript{$\pm$0.99} & 61.69\textsubscript{$\pm$1.51} \\
        \textbf{FedRGL} & 82.12\textsubscript{$\pm$0.51} & \cellcolor[rgb]{ 1,  .902,  .6}78.75\textsubscript{$\pm$0.98} & \cellcolor[rgb]{ 1,  .902,  .6}75.14\textsubscript{$\pm$0.30} & \cellcolor[rgb]{ 1,  .902,  .6}71.52\textsubscript{$\pm$0.14} & \cellcolor[rgb]{ 1,  .902,  .6}66.08\textsubscript{$\pm$1.02} & \cellcolor[rgb]{ 1,  .902,  .6}63.22\textsubscript{$\pm$0.82} & 85.33\textsubscript{$\pm$0.12} & \cellcolor[rgb]{ 1,  .902,  .6}81.77\textsubscript{$\pm$0.32} & \cellcolor[rgb]{ 1,  .902,  .6}76.84\textsubscript{$\pm$0.23} \\
        \bottomrule
      \end{tabular}%
      }
    \end{subtable}
    % \vspace{1em} % Space between tables
    \begin{subtable}{\textwidth}
     \resizebox{\textwidth}{!}{
      \begin{tabular}{c|ccc|ccc|ccc}
        \toprule
        \rowcolor[rgb]{ .749,  .749,  .749} \textbf{Dataset} & \multicolumn{3}{c|}{\textbf{CS (10 Clients)}} & \multicolumn{3}{c|}{\textbf{Photo (20 Clients)}} & \multicolumn{3}{c}{\textbf{Physics (20 Clients)}} \\
       \toprule
       \midrule
        Method & Normal & Uniform & Pair  & Normal & Uniform & Pair  & Normal & Uniform & Pair \\
        \midrule
        \midrule
        FedAvg & 86.95\textsubscript{$\pm$0.09} & 65.95\textsubscript{$\pm$0.21} & 68.95\textsubscript{$\pm$0.60} & 85.88\textsubscript{$\pm$0.18} & 72.55\textsubscript{$\pm$1.15} & 60.52\textsubscript{$\pm$1.49} & 92.78\textsubscript{$\pm$0.19} & 69.05\textsubscript{$\pm$0.26} & 70.96\textsubscript{$\pm$0.98} \\
        FedProx & 88.21\textsubscript{$\pm$0.21} & 65.40\textsubscript{$\pm$0.27} & 68.23\textsubscript{$\pm$0.62} & 86.51\textsubscript{$\pm$0.12} & 69.79\textsubscript{$\pm$1.05} & 62.04\textsubscript{$\pm$0.96} & 93.30\textsubscript{$\pm$0.04} & 69.06\textsubscript{$\pm$0.17} & 71.35\textsubscript{$\pm$0.58} \\
        FGSSL & 88.75\textsubscript{$\pm$0.11} & 69.67\textsubscript{$\pm$0.37} & 71.19\textsubscript{$\pm$1.07} & 83.92\textsubscript{$\pm$0.91} & 70.99\textsubscript{$\pm$1.14} & 64.31\textsubscript{$\pm$1.77} & 93.46\textsubscript{$\pm$0.14} & 78.25\textsubscript{$\pm$0.34} & 76.52\textsubscript{$\pm$0.85} \\
        FedTAD & 86.33\textsubscript{$\pm$0.03} & 72.22\textsubscript{$\pm$0.26} & 70.97\textsubscript{$\pm$1.91} & 85.20\textsubscript{$\pm$1.05} & 70.15\textsubscript{$\pm$1.48} & 55.64\textsubscript{$\pm$6.82} & 93.12\textsubscript{$\pm$0.39} & 70.29\textsubscript{$\pm$0.56} & 70.06\textsubscript{$\pm$0.59} \\
        Co-teaching & 86.78\textsubscript{$\pm$0.08} & 75.90\textsubscript{$\pm$1.38} & 73.14\textsubscript{$\pm$3.14} & 84.52\textsubscript{$\pm$0.25} & 62.99\textsubscript{$\pm$1.11} & 62.96\textsubscript{$\pm$0.47} & 92.71\textsubscript{$\pm$0.13} & 81.67\textsubscript{$\pm$2.68} & 77.74\textsubscript{$\pm$3.56} \\
        FedCorr & 87.49\textsubscript{$\pm$0.14} & 68.37\textsubscript{$\pm$1.01} & 70.14\textsubscript{$\pm$1.26} & 78.42\textsubscript{$\pm$3.32} & 62.62\textsubscript{$\pm$2.39} & 52.13\textsubscript{$\pm$4.89} & 93.19\textsubscript{$\pm$0.26} & 75.36\textsubscript{$\pm$0.63} & 78.57\textsubscript{$\pm$0.73} \\
        FedNoro & 82.87\textsubscript{$\pm$0.18} & 55.15\textsubscript{$\pm$0.44} & 62.42\textsubscript{$\pm$2.99} & 76.82\textsubscript{$\pm$0.01} & 48.51\textsubscript{$\pm$1.32} & 45.15\textsubscript{$\pm$0.34} & 91.36\textsubscript{$\pm$0.69} & 57.65\textsubscript{$\pm$1.10} & 55.50\textsubscript{$\pm$2.36} \\
        FedNed & 87.50\textsubscript{$\pm$0.04} & 66.18\textsubscript{$\pm$0.13} & 66.55\textsubscript{$\pm$0.76} & 83.37\textsubscript{$\pm$0.36} & 66.42\textsubscript{$\pm$1.81} & 63.66\textsubscript{$\pm$1.06} & 92.79\textsubscript{$\pm$0.06} & 68.80\textsubscript{$\pm$0.14} & 72.37\textsubscript{$\pm$0.78} \\
        RHFL  & 87.27\textsubscript{$\pm$0.20} & 56.13\textsubscript{$\pm$0.38} & 56.62\textsubscript{$\pm$1.93} & 83.42\textsubscript{$\pm$0.05} & 66.67\textsubscript{$\pm$1.67} & 66.38\textsubscript{$\pm$2.38} & 93.18\textsubscript{$\pm$0.02} & 58.22\textsubscript{$\pm$0.19} & 65.58\textsubscript{$\pm$1.75} \\
        CRGNN & 87.63\textsubscript{$\pm$0.73} & 81.30\textsubscript{$\pm$0.54} & 74.56\textsubscript{$\pm$1.41} & 85.09\textsubscript{$\pm$0.27} & 71.66\textsubscript{$\pm$1.05} & 66.44\textsubscript{$\pm$1.08} & 92.87\textsubscript{$\pm$0.12} & 83.37\textsubscript{$\pm$0.17} & 80.85\textsubscript{$\pm$0.47} \\
        RTGNN & 84.73\textsubscript{$\pm$0.29} & 69.19\textsubscript{$\pm$0.57} & 70.12\textsubscript{$\pm$0.21} & 82.35\textsubscript{$\pm$0.31} & 73.86\textsubscript{$\pm$1.41} & 67.44\textsubscript{$\pm$0.95} & 91.89\textsubscript{$\pm$0.24} & 74.35\textsubscript{$\pm$3.08} & 75.21\textsubscript{$\pm$1.69} \\
        ERASE & 87.64\textsubscript{$\pm$0.21} & 72.88\textsubscript{$\pm$1.19} & 70.08\textsubscript{$\pm$1.26} & 81.40\textsubscript{$\pm$0.34} & 53.94\textsubscript{$\pm$3.64} & 50.76\textsubscript{$\pm$1.39} & 90.53\textsubscript{$\pm$0.59} & 78.90\textsubscript{$\pm$0.87} & 70.42\textsubscript{$\pm$1.16} \\
        \textbf{FedRGL} & \cellcolor[rgb]{ 1,  .902,  .6}90.58\textsubscript{$\pm$0.13} & \cellcolor[rgb]{ 1,  .902,  .6}87.79\textsubscript{$\pm$0.22} & \cellcolor[rgb]{ 1,  .902,  .6}81.01\textsubscript{$\pm$0.32} & \cellcolor[rgb]{ 1,  .902,  .6}88.42\textsubscript{$\pm$0.27} & \cellcolor[rgb]{ 1,  .902,  .6}84.06\textsubscript{$\pm$0.13} & \cellcolor[rgb]{ 1,  .902,  .6}77.56\textsubscript{$\pm$0.37} & \cellcolor[rgb]{ 1,  .902,  .6}93.62\textsubscript{$\pm$0.11} & \cellcolor[rgb]{ 1,  .902,  .6}90.05\textsubscript{$\pm$0.08} & \cellcolor[rgb]{ 1,  .902,  .6}86.39\textsubscript{$\pm$0.51} \\
        \bottomrule
        \end{tabular}%
        }
       \end{subtable}
       \caption{Comparison of the accuracy of state-of-the-art methods at a noise rate of \(\eta = 0.3\) on six graph datasets, where \textbf{Normal} denotes clean labeling. The best precision is denoted by (    \protect\tikz[baseline=-0.5ex]\protect\node[fill=darkyellow1,rectangle,minimum width=2.0em,minimum height=0.9em]{}; ).}
       \label{tab1}%
\end{table*}\\
\noindent\textbf{Local Predictive Entropy.} To mitigate the spread of erroneous knowledge during global model aggregation, we propose using model predictive entropy to assess each client's model quality. While predictive entropy on a public dataset at the server has been validated in the CV domain \cite{huangself}, \textbf{our approach} does not require an unsupervised public dataset on the server side. Leveraging the transductive training approach in subgraph FL, each client calculates the predictive entropy of its local unlabeled nodes before uploading the model to the server, enabling robust reweighting. Specifically, after the $m$-th client completes the $t$-th training round, the client's model parameters $w_m^t$ are used to compute the predictive entropy $\mathcal{H}_m$ of the unlabeled nodes $V_m^{U}$ (\textit{i.e.,} $V_m^{V_a}$ and $V_m^{T_e}$):
\small
\begin{equation}
\left.\begin{aligned}
\hbar\left(v_m^i\right) &= \frac{-1}{|C|} \sum_{c \in C} \hat{P}_m^{i, c} \log \hat{P}_m^{i, c} \\
\hat{P}_m^i &= \operatorname{Softmax}\left(P_m^i\right)
\end{aligned}\right\} \Rightarrow \mathcal{H}_m=\frac{\displaystyle\sum_{v_m^i \in V_m^{\text{U}}} \hbar\left(v_m^i\right)}{\left|V_m^{\text{U}}\right|}
\phantom{\frac{\sum_{v_m^i \in V_m^{\text{U}}} \hbar\left(v_m^i\right)}{\left|V_m^{\text{U}}\right|}}
\end{equation}
\normalsize
where $\hat{P}_m^{i, c}$ represents the predicted value of class $c$ for the $i$-th node in the unlabeled node set $V_m^{U}$. Next, we explain why the unlabeled nodes $V_m^{U}$ are used to estimate the quality of the model. In noisy label learning, the model becomes increasingly confident in its predictions during the training process, even for noisy nodes \cite{li2023disc}. Therefore, if the labeled nodes are directly used for predictive entropy calculation, it will lead to an unstable estimation of model quality.

\noindent\textbf{Model Aggregation Reweighting}\\
In the model aggregation phase of the $t+1$ communication round, the server collects the model parameters $\{w_i^t\}_{i=1}^M$ and predictive entropies $\{\mathcal{H}_i\}_{i=1}^M$ from the clients. The server then aggregates these models using the predictive entropies to obtain a robust global model $W^{t+1}$:
\begin{equation}
W^{t+1} = \sum_{m=1}^M \frac{H_m}{\sum_{m=1}^M H_m} w_m^t, \quad H_m = \frac{1}{\mathcal{H}_m + \epsilon}
\end{equation}
where $H_m$ is the inverse of the predictive entropy of the $m$-th client (the smaller the predictive entropy, the better the model quality), and $\epsilon$ is a small constant to avoid division by zero, set to $\epsilon = 1e-9$ in this work. Additionally, to ensure a certain degree of reliability for the global model during the initial phase, we conduct $T_{\text{warm}}$ rounds of \textbf{warm-up} training, which only involves standard cross-entropy loss training on clean nodes without noisy node filtering and server-side aggregation reweighting. Due to space limitations, the pseudo-code of the FedRGL algorithm can be viewed in the Appendix. 

\section{Experiment}
\subsection{Experimental Setup}
\noindent\textbf{}\textbf{Datasets.} Our experiments are conducted on six real-world graph datasets, including three citation network datasets (\textit{i.e.,} Cora, CiteSeer, PubMed) \cite{yang2016revisiting}, two co-author datasets (\textit{i.e.,} CS, Physics) \cite{shchur2018pitfalls}, and one user-item dataset (\textit{i.e.,} Photo) \cite{shchur2018pitfalls}. Additionally, to further verify the scalability of our method, we conducted experiments on a large OGB dataset (\textit{i.e.,} ogbn-arxiv) \cite{hu2020open}. \textit{See Appendix for specific explanations.}\

\noindent\textbf{Noise Settings.} In this paper, we assume that the training nodes within each client have label noise with different noise rates. We adopt the same two types of noise settings as in CRGNN \cite{li2024contrastive} (\textit{i.e.,} Uniform and Pair noise). \textit{See Appendix for more detailed explanations.}\

\noindent\textbf{Network Architecture.} Following the settings of the FedTAD method, we adopt a 2-layer GCN as the feature extractor $E$ and the classifier $F$. Additionally, we use a 2-layer MLP as the projection head $I$. The hidden layer size for all datasets is set to be the same as in FedTAD \cite{zhu2024fedtad}.\

\noindent\textbf{Implementation Details.} We follow FedTAD for subgraph partitioning using the Louvain algorithm. All clients use SGD as the optimizer with a weight decay of $5e^{-4}$ and a learning rate of $1 \times 10^{-2}$. Communication rounds are set to 100, and local training epochs to 3. Hyperparameters for FedRGL are tuned using Optuna \cite{akiba2019optuna}, with $\varphi_1$ and $\varphi_2$ in $\{0.5, 1.0, 1.5, 2.0\}$, and $\gamma$ in $\{0.5, 0.6, 0.7, 0.8, 0.9, 0.95\}$. More hyperparameter settings are detailed in the Appendix. Results are reported as mean and standard deviation across 3 random seeds. Experiments were conducted on an NVIDIA A100-PCIE-40GB.

\noindent\textbf{Baseline Methods.} We comprehensively compare FedRGL with global model optimization methods (\textit{i.e.,} FedAvg \cite{mcmahan2017communication}, FedProx \cite{li2020federated}, FGSSL \cite{huang2023federated}, and FedTAD \cite{zhu2024fedtad}), federated label noise learning methods (\textit{i.e.,} FedCorr \cite{xu2022fedcorr}, FedNoro \cite{wu2023fednoro}, FedNed \cite{lu2024federated}, and RHFL \cite{fang2022robust}), as well as graph or image label noise learning methods (\textit{i.e.,} CRGNN \cite{li2024contrastive}, RTGNN \cite{qian2023robust}, Earse \cite{chen2023erase}, and Co-teaching \cite{han2018co}). See Appendix for specific elaboration.
\subsection{Experimental Results}
\noindent\textbf{Generalization Performance.} The classification results of nodes by FedRGL and various state-of-the-art methods under label noise are shown in Tab. \ref{tab1}. As can be seen from the table, our proposed FedRGL achieves the best test accuracy across different datasets and noise types. Moreover, under clean labels (\textit{i.e.,} Normal), our method can match or even surpass existing subgraph FGL methods (\textit{i.e., }FGSSL and FedTAD), demonstrating that FedRGL has superior generalization performance.
\begin{figure}[!t]
    \centering
    \begin{subfigure}[b]{0.5\linewidth}
        \centering
        \includegraphics[width=\linewidth]{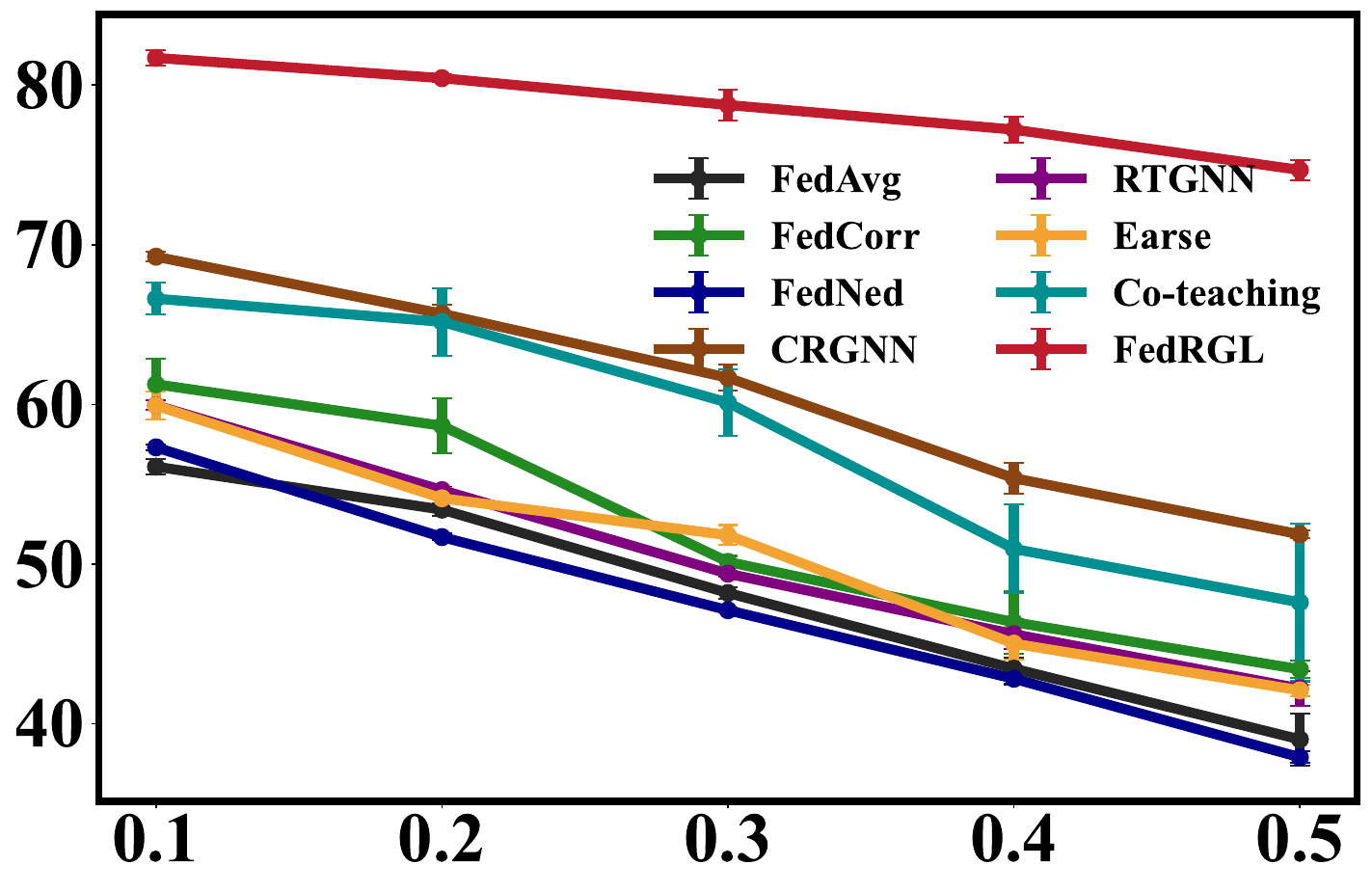} % Replace with your image
        \caption{Uniform Noise \textit{vs.} Noise rate}
        \label{g41}
    \end{subfigure}%
    \hfill
    \begin{subfigure}[b]{0.5\linewidth}
        \centering
        \includegraphics[width=\linewidth]{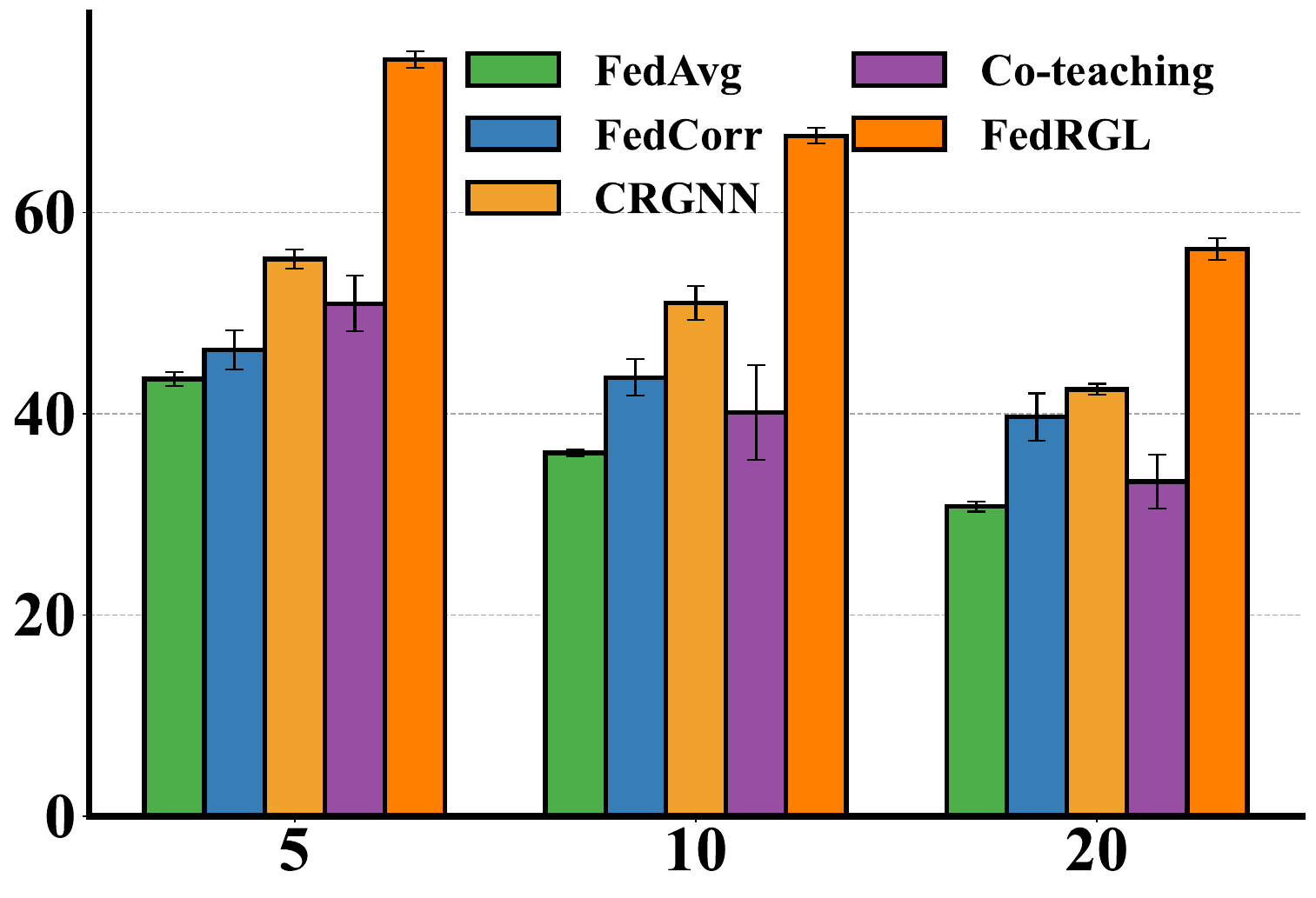} % Replace with your image
        \caption{Pair Noise \textit{vs.} Clients}
        \label{g42}
    \end{subfigure}%
    \caption{Accuracy on Cora under different levels of label noise and numbers of clients.}
    \label{G4}
\end{figure}\\
\noindent\textbf{Different noise scales and client numbers.} We evaluated FedRGL's performance on the Cora dataset under various noise rates and client numbers. As shown in Fig. \ref{g41}, FedRGL consistently outperforms other methods across all noise rates, with the accuracy gap widening as noise increases. Fig. \ref{g42} also illustrates FedRGL's performance across three client numbers with Pair noise on Cora ($\eta=0.4$). FedRGL effectively identifies and corrects noisy nodes, enabling the global model to achieve the highest test accuracy across different client numbers, with an average improvement of over 20\% compared to FedAvg. \textit{Additional validation data is provided in the Appendix.}\\
\begin{table}[!t]
  \centering
  \resizebox{\columnwidth}{!}{  % 使用 \columnwidth 适应单栏宽度
  \begin{tabular}{c|cccccc}
    \toprule
    \rowcolor[rgb]{0.8, 0.8, 0.8} 
    \textbf{Dataset} & \multicolumn{5}{c}{\textbf{Photo (20 Clients)}} \\    
    \midrule
    \midrule
    % \rowcolor[rgb]{0.8, 0.8, 0.8} 
    \textbf{Method} & \textbf{20\%} & \textbf{40\%} & \textbf{60\%} & \textbf{80\%} & \textbf{100\%} \\
    \midrule
    FedAvg & 85.7 \(\pm\) 0.1 & 80.9 \(\pm\) 1.0 & 76.6 \(\pm\) 0.2 & 72.0 \(\pm\) 1.5 & 53.4 \(\pm\) 1.7 \\
    FedCorr & 78.0 \(\pm\) 1.3 & 75.0 \(\pm\) 2.8 & 61.4 \(\pm\) 1.4 & 61.3 \(\pm\) 2.2 & 48.9 \(\pm\) 3.4 \\
    FedNoro & 61.9 \(\pm\) 0.4 & 56.3 \(\pm\) 1.3 & 48.7 \(\pm\) 1.4 & 47.7 \(\pm\) 0.7 & 39.8 \(\pm\) 0.3 \\
    CRGNN & 85.5 \(\pm\) 0.5 & 78.6 \(\pm\) 2.0 & 72.3 \(\pm\) 1.5 & 73.9 \(\pm\) 0.8 & 58.8 \(\pm\) 1.6 \\
    Co-teaching & 84.0 \(\pm\) 0.5 & 77.0 \(\pm\) 0.1 & 66.7 \(\pm\) 0.1 & 70.8 \(\pm\) 0.3 & 56.6 \(\pm\) 0.7 \\
    \textbf{FedRGL} & \cellcolor[rgb]{1, .902, .6}87.5 \(\pm\) 0.1 & \cellcolor[rgb]{1, .902, .6}87.4 \(\pm\) 0.2 & \cellcolor[rgb]{1, .902, .6}82.3 \(\pm\) 0.3 & \cellcolor[rgb]{1, .902, .6}79.1 \(\pm\) 0.1 & \cellcolor[rgb]{1, .902, .6}73.4 \(\pm\) 0.6 \\
    \bottomrule
  \end{tabular}
  }
  \caption{Accuracy on Photo with different number of noisy clients, where the noise type is Pair, \(\eta = 0.4\).}
  \label{tab2}
\end{table}
\begin{table}[!t]
  \centering
  \resizebox{\columnwidth}{!}{  % 使用 \columnwidth 适应单栏宽度
    \begin{tabular}{c|cl|cl}
    \toprule
    \rowcolor[rgb]{0.8, 0.8, 0.8} 
    \multicolumn{1}{c|}{Dataset} & \multicolumn{2}{c|}{obgn-arxiv (20 Clinets)} & \multicolumn{2}{c}{obgn-arxiv (30 Clinets)} \\
    \midrule
    \midrule
    Method & Uniform & \multicolumn{1}{c|}{Pair} & Uniform & \multicolumn{1}{c}{Pair} \\
    \midrule
    FedAvg & 51.83 ± 0.07 & 48.01 ± 0.29 & 50.76 ± 0.21 & \multicolumn{1}{c}{45.33 ± 0.05} \\
    FedCorr & 52.73 ± 0.27 & 46.83 ± 0.55 & 51.59 ± 0.17 & \multicolumn{1}{c}{46.27 ± 0.43} \\
    FedNed & 50.21 ± 0.53 & 48.07 ± 0.27 & 49.47 ± 0.21 & \multicolumn{1}{c}{45.21 ± 0.03} \\
    RTGNN & 52.31 ± 0.05 & 47.33 ± 0.14 & 51.31 ± 0.34 & \multicolumn{1}{c}{44.08 ± 0.17} \\
    CRGNN & OOM   & \multicolumn{1}{c|}{OOM} & 44.42 ± 0.56 & \multicolumn{1}{c}{41.62 ± 0.36} \\
    \textbf{FedRGL} & \cellcolor[rgb]{ 1,  .902,  .6}56.61 ± 0.37 & \cellcolor[rgb]{ 1,  .902,  .6}52.21 ± 0.17 & \cellcolor[rgb]{ 1,  .902,  .6}55.64 ± 0.25 & \cellcolor[rgb]{ 1,  .902,  .6}50.83 ± 0.22 \\
    \bottomrule
    \end{tabular}%
    }
  \caption{Accuracy on the large-scale graph obgn-arxiv.}
  \label{tab3}%
\end{table}\\
\noindent\textbf{Different number of noise clients.} We evaluated the methods on the Photo dataset with varying proportions of noisy clients, as shown in Tab. \ref{tab2}, where $x\%$ indicates the proportion of noisy clients. FedRGL consistently outperforms other methods, including those requiring additional clean and noisy client selection (\textit{e.g.,} FedCorr, FedNoro). Notably, FedRGL's performance advantage increases as the number of noisy clients grows.\\
\noindent\textbf{Large-scale Graph Data Results.} To evaluate the performance of FedRGL on large-scale graph data, we conducted experiments on the \textit{obgn-arxiv} dataset (\(\eta = 0.4\)). The statistical results are shown in Tab. \ref{tab3}. The current federated learning methods almost fail to improve the impact of noisy labels on large-scale graph datasets, whereas our FedRGL method can improve accuracy by at least 4\% compared to the baseline method \textbf{FedAvg}.
\subsection{Diagnostic Analysis}
\begin{figure}[!t]
    \centering
    \begin{subfigure}[b]{0.5\linewidth}
        \centering
        \includegraphics[width=\linewidth]{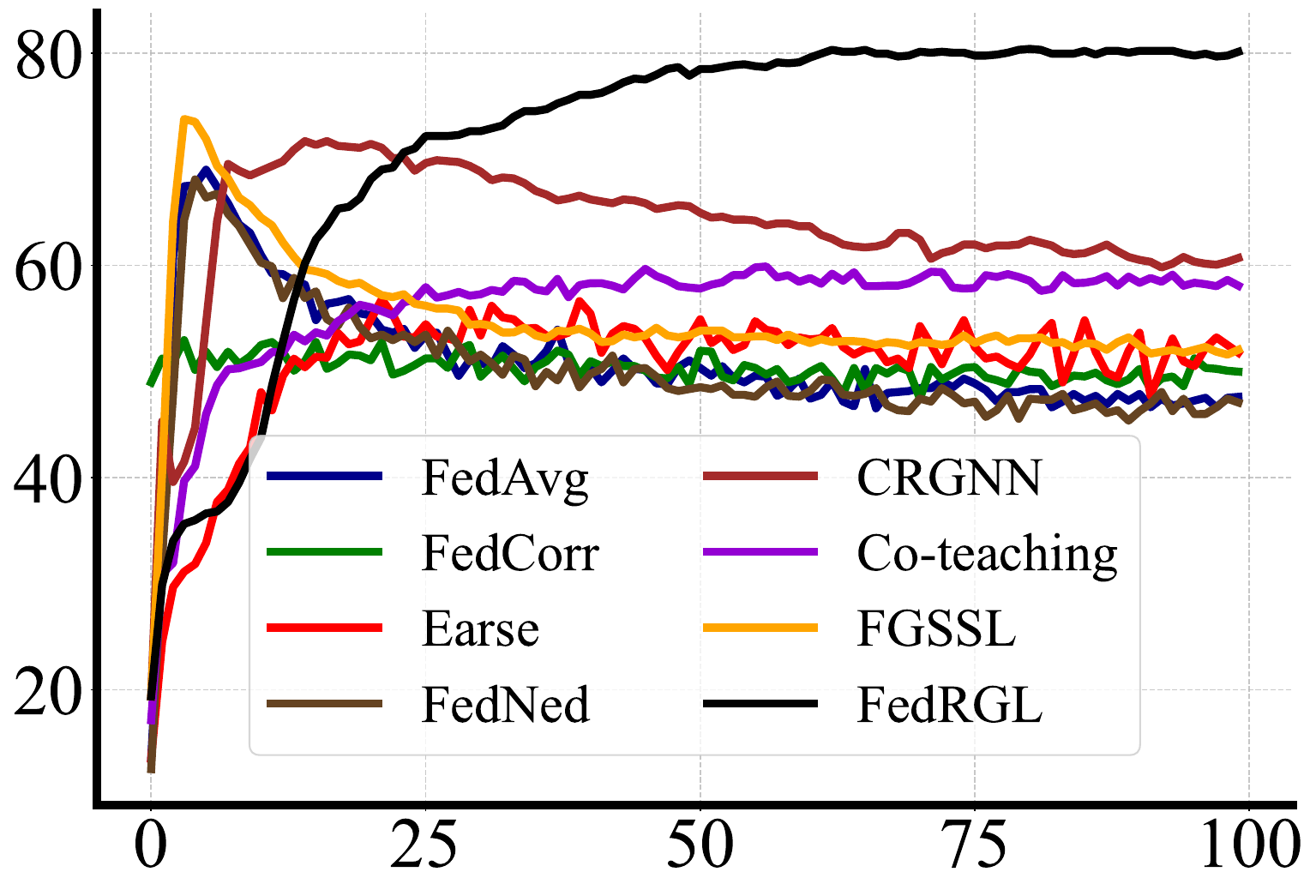} % Replace with your image
        \caption{Cora \textit{vs.} 5 Clients}
        \label{g61}
    \end{subfigure}%
    \hfill
    \begin{subfigure}[b]{0.5\linewidth}
        \centering
        \includegraphics[width=\linewidth]{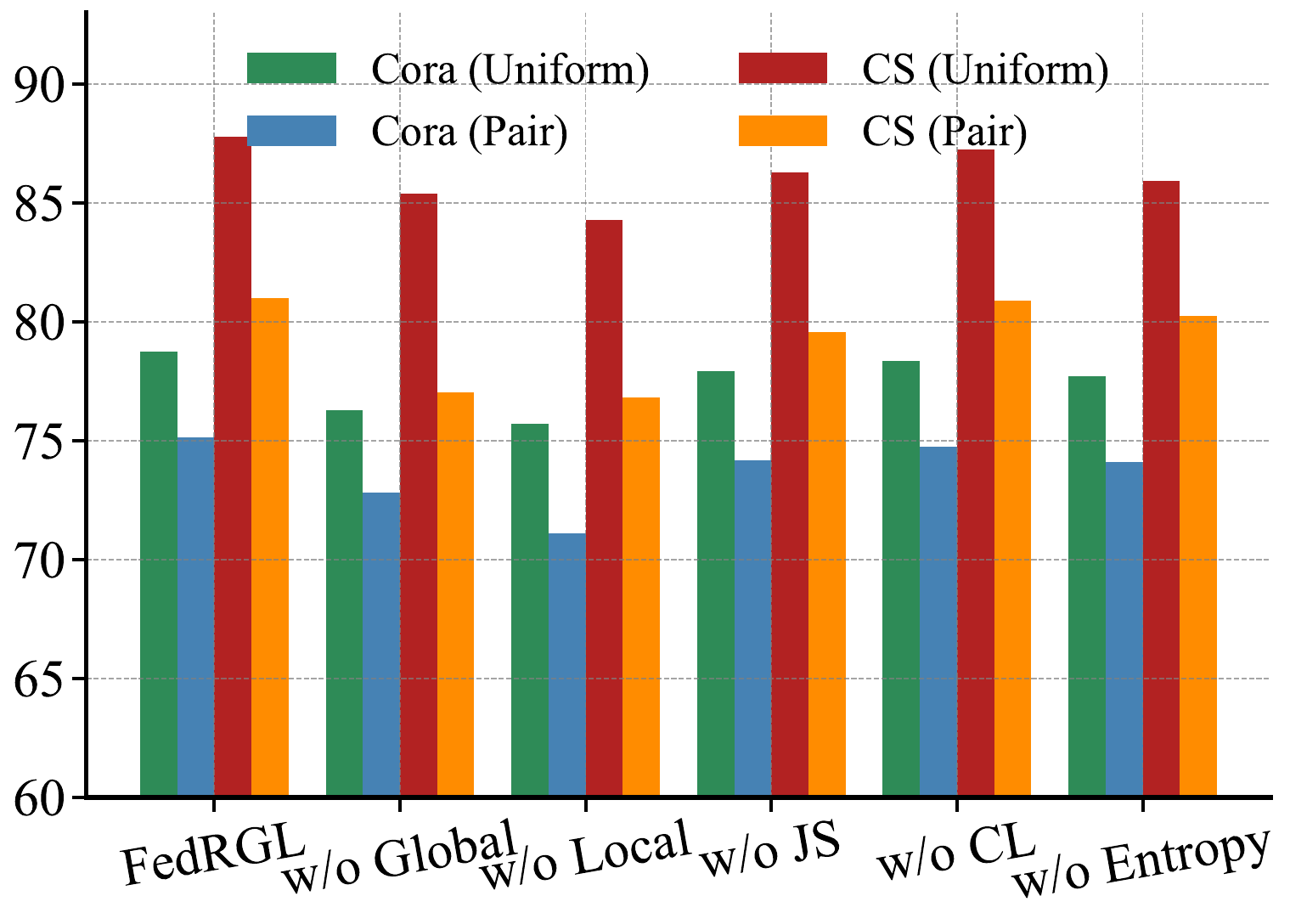} % Replace with your image
        \caption{Ablation experiment}
        \label{g62}
    \end{subfigure}
    \caption{Visualization of the training curves and ablation experiment.}
    \label{G6}
\end{figure}

\noindent\textbf{Stability Analysis.} Fig. \ref{g61} shows the visualization of the training curves under Uniform label noise with $\eta = 0.3$ on Cora. It can be observed that the training stability of FedRGL is consistently better than other methods under label noise, and unlike CRGNN, it does not exhibit a trend of increasing accuracy followed by a decline.
\begin{figure}[!t]
    \centering
    \begin{subfigure}[b]{0.5\linewidth}
        \centering
        \includegraphics[width=\linewidth]{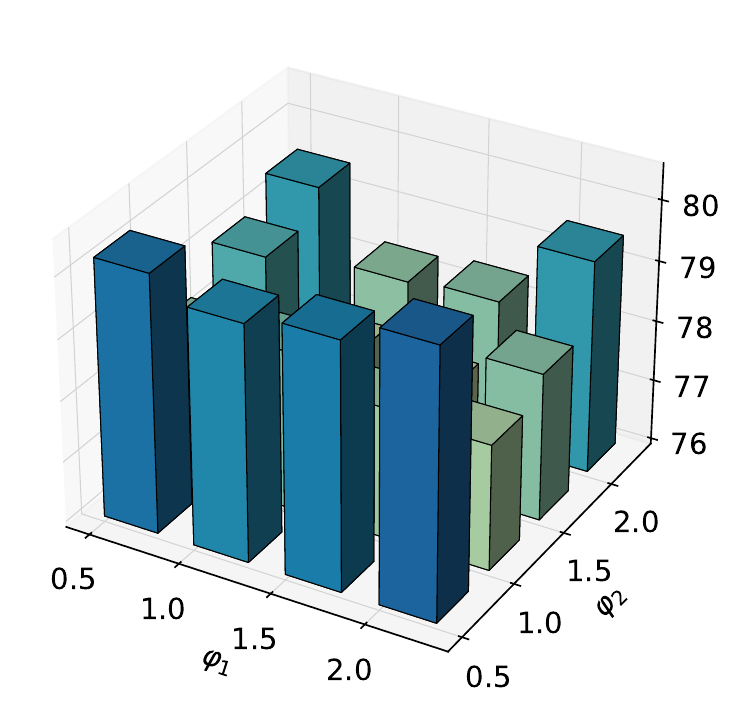} % Replace with your image
        \caption{Loss parameter $\varphi_1$, $\varphi_2$ }
        \label{g71}
    \end{subfigure}%
    \hfill
    \begin{subfigure}[b]{0.5\linewidth}
        \centering
        \includegraphics[width=\linewidth]{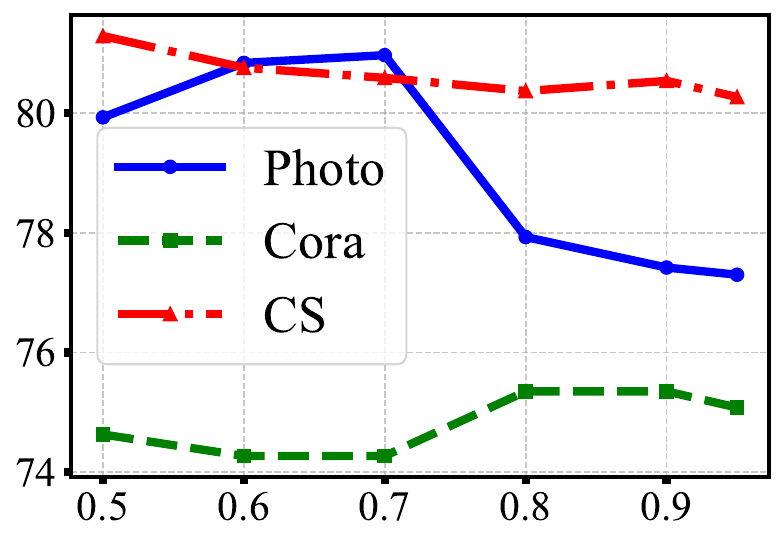} % Replace with your image
        \caption{Confidence parameter $\gamma$}
        \label{g72}
    \end{subfigure}
    \caption{Analysis on hyper-parameter in FedRGL.}
    \label{G7}
\end{figure}\\
\noindent\textbf{Ablation Experiment.} We conducted ablation experiments on the Cora and CS datasets with $\eta=0.3$ to assess the impact of different FedRGL components. Fig. \ref{g62} lists the results for FedRGL and its variants: removing the global model view, local structure view, JS divergence, contrastive learning, and server reweighting. The global and local views are essential for noise filtering, while removing contrastive learning has minimal effect. Both JS divergence and predictive entropy weighting improve performance. \textit{More validation can be found in the Appendix.}\\
\noindent\textbf{Hyperparameter Sensitivity Analysis.} Fig. \ref{G7} examines the dynamic loss in relation to hyperparameters $\varphi_1$, $\varphi_2$, and confidence parameter $\gamma$. In Fig. \ref{g71}, results on Cora with uniform noise rate 0.3 show that $\varphi_1$ (global model view) is relatively insensitive, while $\varphi_2$ (local structural view) shows fluctuations with different settings. Generally, combining $\varphi_1$ and $\varphi_2$ improves model performance, and setting $\varphi_1 = \varphi_2$ has minimal impact on performance. Fig. \ref{g72} illustrates results on Cora, Photo, and CS with a pair noise rate of 0.3, showing small sensitivity fluctuations in Cora and CS, while a large $\gamma$ in Photo reduces the number of high-confidence pseudo-labels. \textit{Further hyperparameter analysis is in the appendix}.

\section{Conclusion}
This paper addresses the challenge of label noise in subgraph FL, distinguishing it from existing federated label noise learning approaches. To mitigate the impact of label noise on the global model in subgraph FL, we propose FedRGL. This method employs dynamic dual-consistency filtering with pseudo-label augmentation on the client side, combined with predictive entropy-based reweighting on the server side, to achieve a noise-robust global model. Experimental results demonstrate that FedRGL consistently outperforms existing methods.

% \clearpage % 强制分页

% \bibliography{aaai25}

\bibliography{aaai25}

\end{document}